\documentclass[11pt]{article}
     \usepackage[preprint]{neurips_2020}

\usepackage{natbib}
\setcitestyle{numbers,square}
\usepackage[utf8]{inputenc} 
\usepackage[T1]{fontenc}    
\usepackage{hyperref}       
\usepackage{url}            
\usepackage{booktabs}       
\usepackage{amsfonts}       
\usepackage{nicefrac}       
\usepackage{microtype}      

\usepackage{graphicx}    
\usepackage{amsmath}
\usepackage{amsfonts}
\usepackage{amssymb}
\usepackage{bbm}
\usepackage{algorithm}
\usepackage{algorithmic}

\DeclareMathOperator*{\argmax}{arg\,max}

\title{A mixed policy to improve performance of language models on math problems}

\author{%
  Gang Chen
    \\
  Vividity tech \\
  \texttt{info@vividitytech.com} \\
}

\begin{document}         
\date{}

\maketitle
\begin{abstract}
When to solve math problems, most language models take a sampling strategy to predict next word according conditional probabilities. In the math reasoning step, it may generate wrong answer. Considering math problems are deterministic, we propose a mixed policy exploration approach to solve math problems with reinforcement learning. In peculiar, we propose a two level token exploration policy: the abstract level explores next token with probability and the second level is deterministic. Specifically, the abstract level policy will decide whether the token is operator or operand with probability sampling, while the second level is deterministic to select next token with the highest score in a greedy way. We test our method on GSM8K dataset with GPT-2 model, and demonstrate more than $2\%$ performance gain. Our implementation is available at https://github.com/vividitytech/math\_lm\_rl.
\end{abstract}

\section{Introduction}

The grad math problems always have the deterministic solutions. That means that given the same question multiple times, our solvers should generate the same results, not matter using language models or rule-based approaches. In other words, if we use language models to solve the math problems, it would be better to take a deterministic approach. However, the language models sample a completion according to conditional distributions, and it may generate different answers while running multiple times. What is more, it may generate wrong answer. That is why at test time, some approaches need to train a verifier, and then generate many candidate solutions and select the one ranked highest by the verifier \cite{Cobbe21}. 

It is challenge to leverage language models to solve the math problems because the former is generative model with sampling while the latter is deterministic. Recent language models \cite{Ziegler2020,Cobbe21} train another reward model to rank generated candidates and select the highest one. This requires large high quality labeled data in either reasoning step or outcome-based supervision \cite{Uesato2022,Lightman2023lets}. Although these methods have required the model to align with human feedback by fine-tuning, it is still challenge to generate desired results. What is even worse it may produce wrong answer if one word (or token) is mistaken for another. The reason behind that is language models generate sequence with probabilities which are learnt statistically from data.  For example  the word "divide" can be followed by "divide into", "divide by" and so on. In the math algebra, "divide into" indicates multiplication operator "*", while "divide by" in most case means division "/". While predicting the next token with language model, if it mistake one for another i.e. "*" for "/", then it generates the wrong sequence, and furthermore the wrong answer to the math problem. 
In this paper, we want to correct this kind of error with RL. On the one hand, we need policy exploration with random sampling; on the other hand, we need deterministic approach to generate answers for math problems. Specifically, we want to explore the operators as randomly as possible while taking a deterministic policy on non operators. 

Considering math problems are deterministic, we propose a mixed policy exploration approach to solve math problems with reinforcement learning. 
Given the high cost of human feedback, we focus on the outcome-based approach and use the answer to extract rewards to improve the performance.  We discount the reward from the final step to each previous step to guide the reasoning process with RL, and propose a two level policy exploration approach: the abstract level policy will decide whether the token is operator or operand and sample from operators, and the second level will take the greedy token selection for non-operators. We test our method on GSM8K dataset with GPT-2 model, and demonstrate more than $2\%$ performance gain.

\section{Related work}
Recent progress in language models has sparked significant interest to solve Math problems. Cobbe et al. released the multi-step mathematical reasoning dataset GSM8K \cite{Cobbe21} and trained verifiers to solve this Math word problem. In test, this method sampled multiple candidates for each question, so the verifier was used to select the highest ranked one as the answer. 
Another large language models (LLMs) called Minerva \cite{lewkowycz2022solving} was proposed to solve mathematical and scientific questions using several techniques, such as few-shot prompting, chain of thought and majority voting, to improve the performance on STEM reasoning tasks. Recently, Symbolic Solvers \cite{heyueya2023solving} combined LLMs with external symbolic solver to perform complex reasoning and calculation, and shown promising results for solving math word problems. 

Language models are generative, which generate next word with conditional probability learnt from training data and may generate wrong answer. It is much better if human can provide feedback to correct the mistake. Reinforcement learning from human preferences (RLHF) have shown popularity which can effectively solve complex RL tasks without access to the reward function \cite{Christiano2017}. In this work \cite{Christiano2017}, users can provide preferences between pairs of trajectory segments, and such feedback can be used to learn better policy. Ziegler et al. fine-tune the pretrained language models from human preferences \cite{Ziegler2020} to improve performance on four NLP tasks. InstructGPT \cite{Ouyang2022} also used RLHF approach to align language models with user intent on a wide range of tasks.

More recently, researchers have shown that the granularity of human feedback \cite{Uesato2022,Lightman2023lets} also matters while fine-tune LLMs.
For example, the outcome-based approaches only need supervision from human on the final result, while process-based approaches require to supervise each reasoning process. Recent language models have yielded significant progress on math problems with RL, either outcome-based supervision or  process supervision. Uesato et al. compared outcome-based approaches and process-based approaches \cite{Uesato2022} and yield many interesting results. 
Lightman et al. conduct the investigation \cite{Lightman2023lets} and find process supervision significantly outperforms outcome supervision for training models to solve problems from the challenging MATH dataset.
In technical level, process supervision requires human feedback on each reasoning step, which is labor intensive and time consuming. Given the high cost of human feedback, we prefer the final answer feedback, which is much efficient for human in the loop approach. In addition, when to solve math problems, most language models take a sampling strategy to predict next operator such as "+", "-", "*" and "/". If it mistakes "*" for "/" in the math reasoning step, it will generate wrong answer. That means we need a deterministic approach to generate answer but also explore possible mistakes to improve the robustness of the language models. In this paper, we propose to learn language models with a two level mixed policy exploration strategy, and try to solve the math problems in a deterministic manner. 

\section{Model}
We present the mixed policy to fine-tune language models. In the following parts, we will introduce the language models and then discuss reinforcement learning with mixed policy to fine-tune model parameters.
\subsection{Background}
Given a vocabulary $\mathcal{V}$ and an ordered sequence of symbols (or tokens) $(x_1,x_2, ..., x_{n})$ with $x_i  \in \mathcal{V}$,  the language model \cite{Bengio2003} is defined as the joint probability over sequences of tokens $\mathcal{V}^n$, which is factorized into product of conditional probabilities
\begin{align}\label{eq:lm}
p(x_1,x_2, ..., x_{n}; \theta) = \prod_{1 \le i <n} p(x_i| x_1, x_2,..., x_{i-1}; \theta)
\end{align}
where the vocabulary $\mathcal{V}$ is a large but finite set, and $\theta$ is the model parameter. $p(x_i | x_1, x_2,..., x_{i-1})$ is conditional probability to predict next word given the previous sequences.

Many NLP problems can be formulated as $p(Y|X; \theta)$, where $X \in \mathcal{V}^n$ is the input sequence and $Y \in \mathcal{V}^m$ is the output.   
There have been many models that can compute these conditional probabilities, such as recurrent neural networks LSTM \cite{HochSchm97} and self-attention Transformer \cite{Vaswani2017attention}. Especially the transformer architecture have significant improvements in the expressiveness and accuracy of models \cite{Radford2018,Brown2020}. To learn the model parameters $\theta$, we can use cross entropy loss: 
\begin{align}\label{eq:loss1}
L_i(\theta)  = - \log p(y_i)  =  - \log p(y_i | y_{<i}, X; \theta) 
\end{align}
where only $y_i$ holds and other tokens in $\mathcal{V} \backslash y_i$ are zeros.
Then, the cross entropy error over the sequence of size $m$ is:
\begin{align}\label{eq:loss}
 L(\theta) = \sum_{i=1}^m L_i(\theta) = - \sum_{i=1}^m  \log p(y_i) 
\end{align}

Given the pertained model such as GPT-2, the next step is to fine-tune it to perform well on math problem, such as GSM8K dataset \cite{Cobbe21}.

The conditional probability $p(y_i) = p(y_i | y_{<i}, X; \theta)$ to predict next token can be thought as the action exploration in reinforcement learning (RL). If we know the final outcome, then the reward can be used as feedback to fine-tune model parameter to guide the sequence generation. The whole process matches well with the objective of reinforcement learning, which is to maximize a cumulative return with sequential interactions between an agent and its environment \cite{SuttonB98,Williams92}. We initialize a policy $\pi=p$, and then fine-tune $\pi (\theta)$ to perform the task well using RL. If the task was defined by a reward function $r: X\times Y\rightarrow \mathbb{R}$, then we could use RL to directly optimize the expected reward:
\begin{align}\label{eq:objj}
J(\theta) = E_{\tau \sim \pi_{\theta}(\tau)} [  \sum_{i=1}^T  \gamma^t r(  X, \hat{y_i}   )   ] = E_{\tau \sim \pi_{\theta}(\tau)} [  R_1^T ]   
\end{align}
where $\tau$ is the trajectory $\tau =\hat{Y}= \{  \hat{y_i}  \}_{i=1}^T$ generated while following the policy $\pi (\theta)$. Note that while predicting the next symbol $\hat{y}_i \sim p(y_i | y_{<i}, X; \theta)$, we can either sample it or take a greedy strategy to select $\hat{y}_i$ with maximum probability. Recently, both process supervision and output supervision \cite{Uesato2022,Lightman2023lets} have been proposed to define the reward $r: X\times Y\rightarrow \mathbb{R}$ to improve the performance on math problems. Considering the high cost for each step supervision, we focus on the final outcome-based approach. And we define $r(X,\hat{y}_T)=1$ if the final result $\hat{y}_T$ in $\hat{Y}$ matches its ground truth from $Y$. Note that we do not require the sequence generated $\hat{Y}$ from $\pi(\theta)$ match $Y$ in either sequence length or symbol level.

\subsection{Objective function}
In the fine-tune stage, we require the generated $\hat{Y}$ matches the ground truth $Y$ and also gives the right answer from RL policy update.
\begin{align}\label{eq:cerl}
loss & = L(\theta) + \alpha J(\theta)  \nonumber \\
        & =  -\sum_i \log p(y_i)  +\lambda \max \big(0,   - E_{\tau \sim \pi_{\theta}(\tau)}   \log  \pi_{\theta}(\tau)   ( R_1^T  -b)    \big)
\end{align}
where $\lambda$ is the weight to balance two items, $R_1^T$ are the rewards over sequence (which discounts the reward from the final step to each previous step) and $b$ is the baseline which can be approximated by neural network. For the REINFORCE loss, we add maximum to suppress the policy lower than average. In this work, we add additional linear layer over the hidden outputs from GPT-2 to model $b$. Note that the unequal length between $\hat{Y}$ and $Y$ can be addressed with padding to match $Y$ in the training stage.

In this paper, we introduce another action exploration schema, and add another abstract level over the current policy. In this abstract level, we categorize the tokens into 5 classes: 4 operators and the rest from $\mathcal{V}$, where operators can be "+", "-", "*" and "/". We use $\mathcal{\underline{V}}$  to denote all rest tokens except operators, and $\mathcal{O}=\{+, -, *, /, \mathcal{\underline{V}} \}$ to denote the action space for convenience. And we have $|\mathcal{O}|=5$ and we also need to build bidirectional mapping between $\mathcal{V}$ and $\mathcal{O}$. Suppose we use $f: \mathcal{V} \rightarrow \mathcal{O}$, then $o_i = f(y_i)$, for $i \in [1, m]$. And there is inverse mapping $y_i = f^{-1}(o_i)$ if $o_i$ is from the operators.

In the abstract level, we build a linear model to model the distribution over $\mathcal{O}$. Its input is the hidden output from transformer and its output is the action space $\mathcal{O}$ the policy can take. Thus, we can define the abstract level policy as:
\begin{align}\label{eq:abspolicy}
p(o_i) = p(o_i | h_{<i}, X; \theta_o) = \text{softmax} (h_{<i} \theta_o)
\end{align}
where ${o}_i$ is the next predicted token given the previous context, $h_{<i}$ is the last hidden output and $\theta_o$ is the weight mapping from the hidden output $h_{<i}$ to $\mathcal{O}$. 

Then we define the new mixed policy as follows:
\begin{align}\label{eq:nexttoken}
\hat{y}_i   = \begin{cases} f^{-1}(\hat{o}_i), with\ sample \ \hat{o}_i \ from \ p(o_i) & \text{if} \  \hat{o}_i  \text{ is the operator} \\
                \argmax_{y_i} p(y_i)  & \text{otherwise} \end{cases}
\end{align}
where $\hat{o}_i$ is sampled according to Eq. \ref{eq:abspolicy}.

In order to learn $\theta_o$, we need to map $ y_i \in \mathcal{V}$ into $o_i \in \mathcal{O}$. In other words, for any $(X, Y)$ pair, we need to cast $Y$ from space $\mathcal{V}$ into $\mathcal{O}$. Then for each $y_i \in Y$, we have $o_i \in Y^{\mathcal{O}}$ as the corresponding label. 
And our new loss function is:
\begin{align}\label{eq:floss}
loss  =  -\sum_i \log p(y_i) - \beta \sum_i \log p(o_i) + \lambda  \max \big(0, - E_{\tau \sim \pi_{\theta}(\tau)}   \textrm{log}  \pi_{\theta}(\tau)   ( R_1^T  -b)    \big)
\end{align}
where $\beta$ and $\lambda$ are constants to balance the three items above. Note that the loss is only for one sequence and we can sum it to handle batch processing in both training and inference.

\subsection{Algorithm}
We summarize our approach in Algorithm. \ref{alg:algd2q}. The basic idea is that given the current question $X$ and the model parameter $\theta$, we can generate sequence $\hat{Y}$ using our mixed policy in Eq. \ref{eq:nexttoken}. Then we compare $\hat{Y}$ with its groundtruth $Y$ to decide the reward $r$. Then we fine-tune the model parameter $\theta$ from the reward $r$.

\begin{algorithm}[tb]
\caption{Training}
\label{alg:algd2q}
\begin{algorithmic}
\STATE Initialize neural networks and its parameters from GPT
\FOR{epoch = 1 {\bfseries to} $N$}
\STATE sample $(X, Y)$ from training data 
\STATE $\hat{ Y} = [ \  \ ]$ // generate sequence given $X$
\FOR{$i = 1$ {\bfseries to} $T$}
\STATE sample $o_i$ according to Eq. \ref{eq:abspolicy}
\STATE generate next token $\hat{y}_i$ according to Eq. \ref{eq:nexttoken}
\STATE $\hat{ Y} =[\hat{ Y} ,  \hat{y}_i]$ 
\ENDFOR
\STATE compare $(X, Y)$ and $(X, \hat{Y})$, then compute the reward $r$
\STATE take gradient step to minimize Eq. \ref{eq:floss}
\ENDFOR
\STATE Return model and parameters.
\end{algorithmic}
\end{algorithm}

In the inference stage, given the question, we take the maximum $\hat{y}_i  = \argmax_{y_i} p(y_i) $ for each $i$ to generate the answer.

\section{Experimental results}

We used the smallest version of GPT-2 \cite{Vaswani2017attention,Radford2018}, with 124M parameters to evaluate our algorithm. Given the same model, we want to estimate how our algorithm improves the model performance. In addition, for large models, it is hard to judge whether the performance gain is from model size or the algorithm itself.

We did the experiments on the GSM8K grad math dataset \cite{Cobbe21}. We chose GSM8K because it provides the right answer for each question, so that we can exact rewards to improve policy. 
Similar to \cite{Uesato2022}, we split out the original training set with 7118 examples for training and the rest 256 examples for validation, and keep 1319 test examples for testing. We used the default Adam algorithm with learning rate =$1e^{-5}$. As for parameter setting, we used batch size =1, $\beta=1$ and $\lambda=0.1$ for all experiments.

\begin{figure}[t!]
\begin{tabular}{c}
\includegraphics[trim=5mm 20.0mm 5mm 25mm, clip,  width=14.2cm]{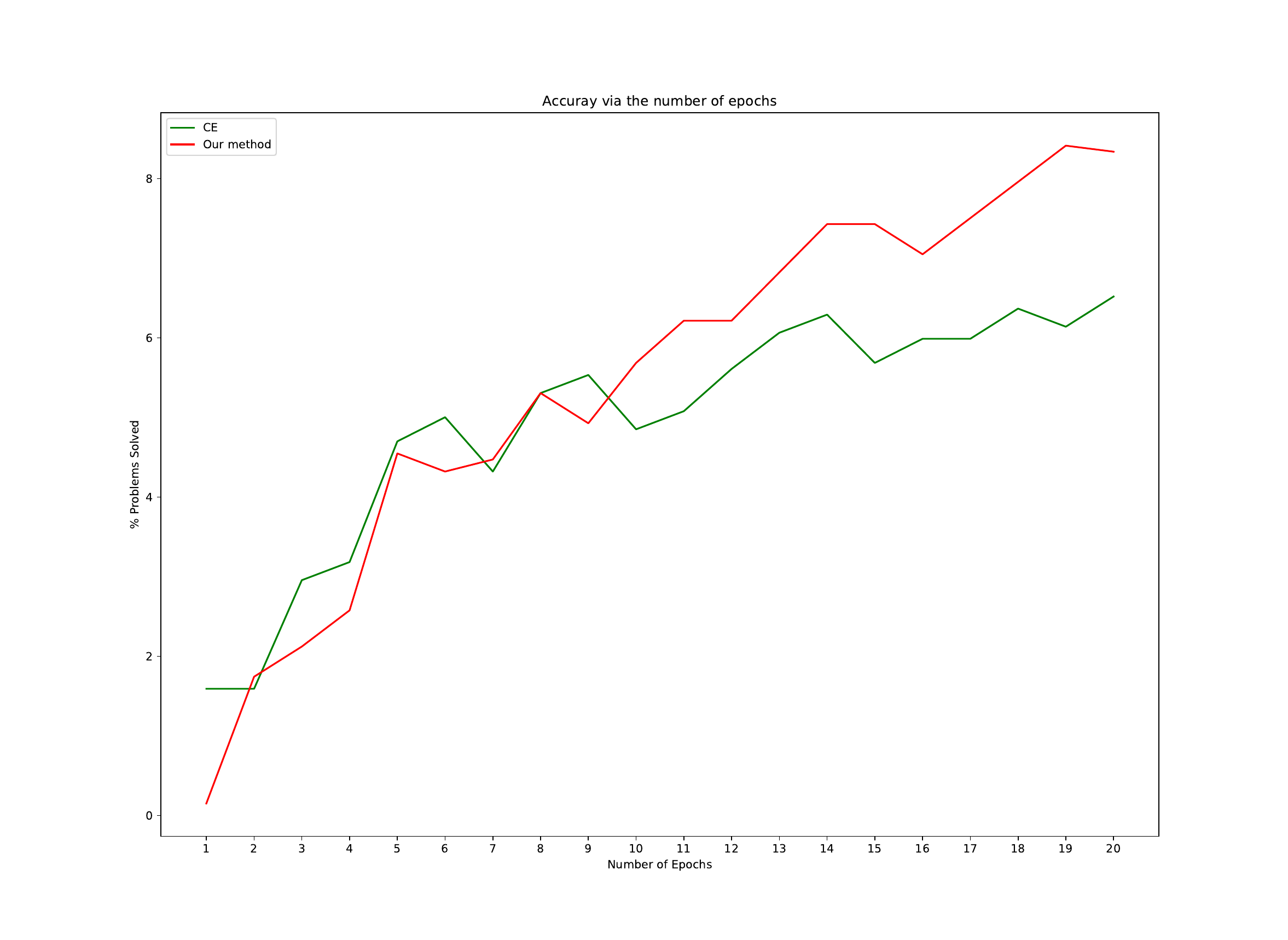}
\end{tabular}
\caption{The figure shows how the accuracy varies with the number of epochs. Our method yields higher accuracy than cross entropy loss (CE).}
\label{Fig:rlcomp}
\end{figure}

We tested different algorithms cross entropy loss (CE), CE+RL (Eq. \ref{eq:cerl}) and our algorithm (Eq. \ref{eq:floss}). The result in Table \ref{tab:tab1} shows that our method with RL can boost performance more than $2\%$ in 20 epochs training. We also compare our approach to CE in Fig. \ref{Fig:rlcomp}, and demonstrate that our method gain significantly after 10 epochs. 

\begin{table}[h]
\centering
\begin{tabular}{ |c|c|c|c| } 
\hline
\multicolumn{4}{|c|}{\% problems solved}\\
\hline
&  CE & CE+RL & Our method \\ \hline
 GSM8K & 6.4\% &  8.11\% & 8.42\%   \\ \hline
\end{tabular}
\caption{Accuracy}
\label{tab:tab1}
\end{table}

\begin{figure*}[h!]\centering
\begin{tabular}{c}
\includegraphics[trim=5mm 40.0mm 5mm 50mm, clip,  width=12.2cm]{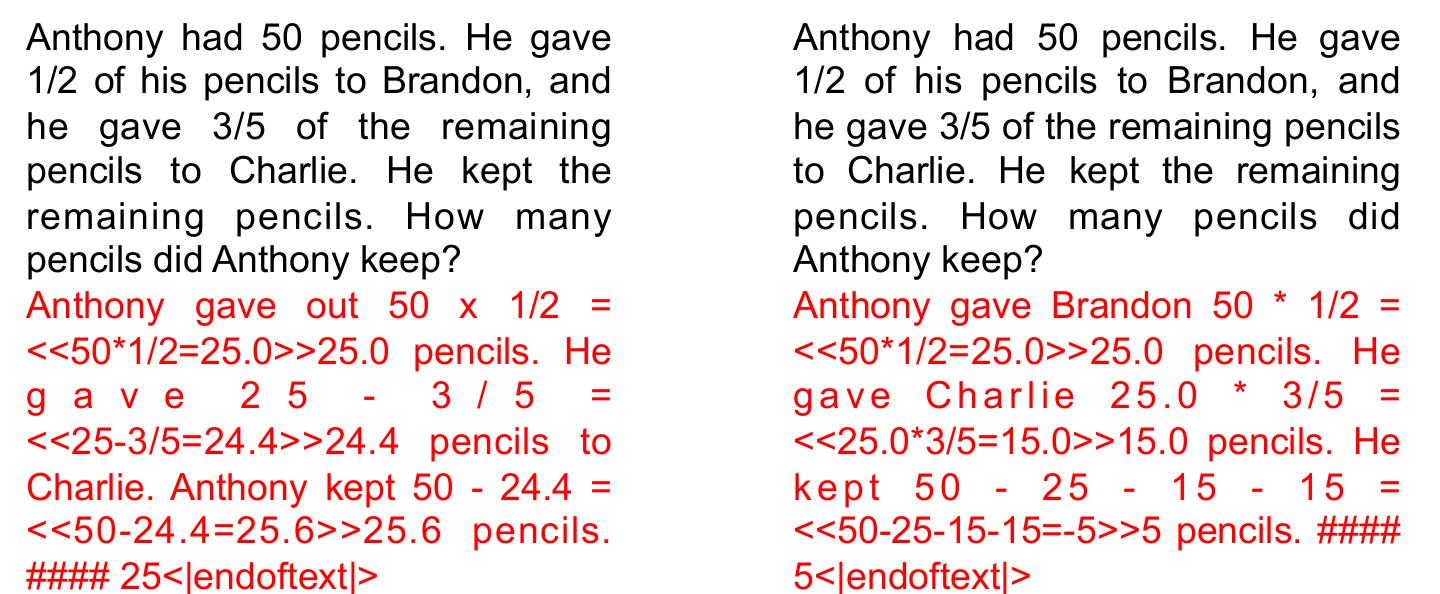}
\end{tabular}
\caption{The wrong results from cross entropy loss (CE) and our approach. It indicates that the GPT-2 model cannot understand the problem well. And it is possible to understand semantic meaning better by increasing the model size.}
\label{Fig:bothwrong}
\end{figure*}

\begin{figure*}[h!]
\begin{tabular}{c}
\includegraphics[trim=0mm 20.0mm 0mm 20mm, clip,  width=13.2cm]{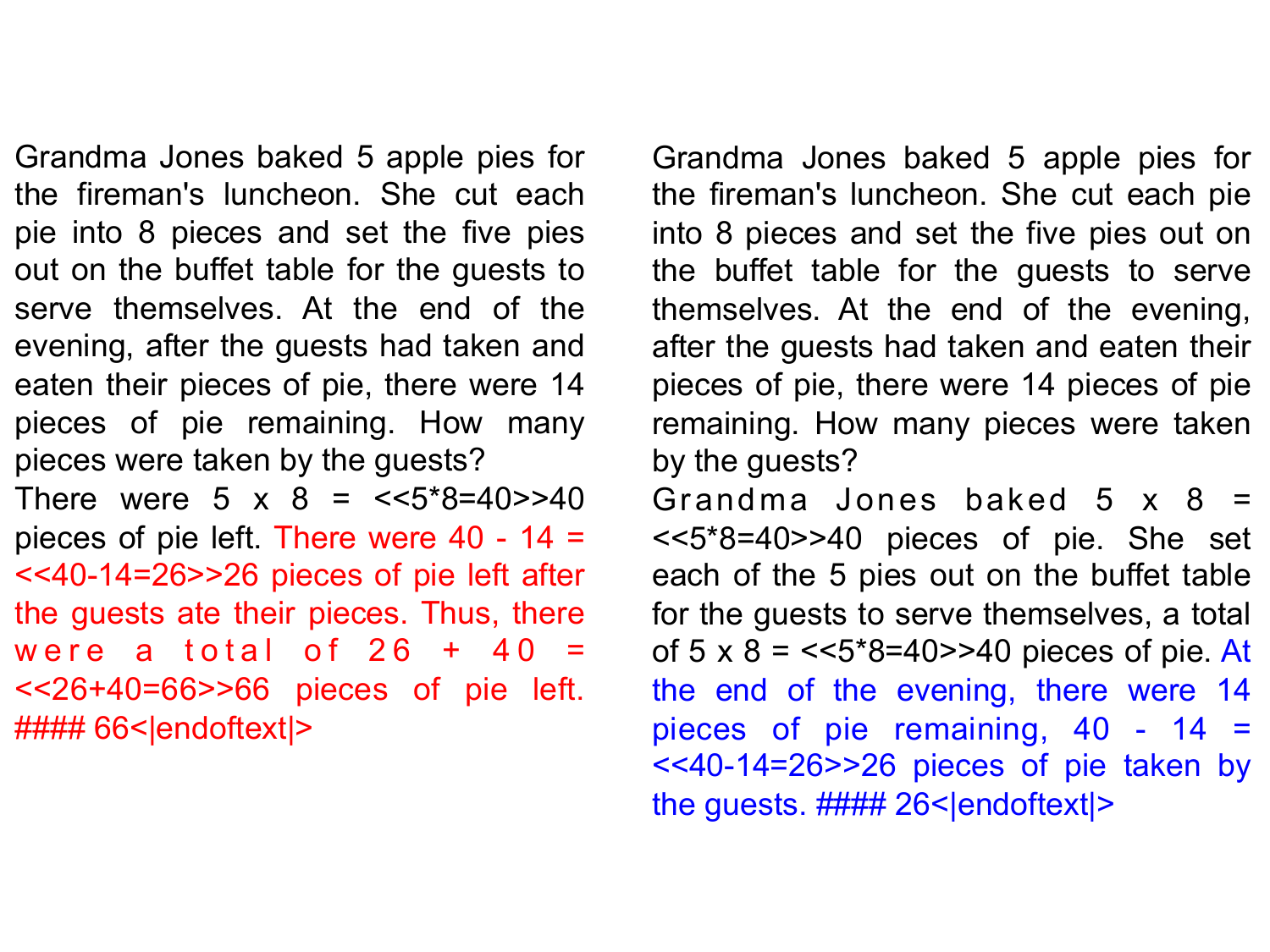}
\end{tabular}
\caption{The left is the result from cross entropy loss (CE), and the right is from our approach. It shows that our model can generate correct answer.}
\label{Fig:goodone}
\end{figure*}

We also listed some examples generated from CE and our method. Figure \ref{Fig:bothwrong} shows the wrong results from both CE and our algorithm. It seems that the GPT model cannot understand the semantic meaning of the given question. It is possible to solve it with a large size model. Another example in Figure \ref{Fig:goodone} shows that our model can correct the mistake from CE.

\section{Conclusion}
In this paper, we propose a mixed policy exploration approach to solve math problems with reinforcement learning. Specifically, we explore the operators to trade off random and deterministic in solving the math problems, so that we can always generate the same deterministic results. The experimental analysis shows that our model can boost the performance of language models on the GSM8K dataset.

Because our approach only explore operators, it may generate wrong operands in RL. In the future work, we will consider to explore both operators and operands to boost the robustness of the model. Also we can improve the logic understanding of the math problems by increasing the model size, and then correct the logic mistake with RL to improve the performance.
%
%


\bibliographystyle{unsrt}
\bibliography{llm2022}

\begin{thebibliography}{10}

\bibitem{Cobbe21}
Karl Cobbe, Vineet Kosaraju, Mohammad Bavarian, Mark Chen, Heewoo Jun, Lukasz
  Kaiser, Matthias Plappert, Jerry Tworek, Jacob Hilton, Reiichiro Nakano,
  Christopher Hesse, and John Schulman.
\newblock Training verifiers to solve math word problems.
\newblock {\em CoRR}, abs/2110.14168, 2021.

\bibitem{lewkowycz2022solving}
Aitor Lewkowycz, Anders Andreassen, David Dohan, Ethan Dyer, Henryk
  Michalewski, Vinay Ramasesh, Ambrose Slone, Cem Anil, Imanol Schlag, Theo
  Gutman-Solo, Yuhuai Wu, Behnam Neyshabur, Guy Gur-Ari, and Vedant Misra.
\newblock Solving quantitative reasoning problems with language models, 2022.

\bibitem{heyueya2023solving}
Joy He-Yueya, Gabriel Poesia, Rose~E. Wang, and Noah~D. Goodman.
\newblock Solving math word problems by combining language models with symbolic
  solvers, 2023.

\bibitem{Ouyang2022}
Long Ouyang, Jeff Wu, Xu~Jiang, Diogo Almeida, Carroll~L. Wainwright, Pamela
  Mishkin, Chong Zhang, Sandhini Agarwal, Katarina Slama, Alex Ray, John
  Schulman, Jacob Hilton, Fraser Kelton, Luke Miller, Maddie Simens, Amanda
  Askell, Peter Welinder, Paul Christiano, Jan Leike, and Ryan Lowe.
\newblock Training language models to follow instructions with human feedback,
  2022.

\bibitem{Uesato2022}
Jonathan Uesato, Nate Kushman, Ramana Kumar, Francis Song, Noah Siegel, Lisa
  Wang, Antonia Creswell, Geoffrey Irving, and Irina Higgins.
\newblock Solving math word problems with process- and outcome-based feedback,
  2022.

\bibitem{Lightman2023lets}
Hunter Lightman, Vineet Kosaraju, Yura Burda, Harri Edwards, Bowen Baker, Teddy
  Lee, Jan Leike, John Schulman, Ilya Sutskever, and Karl Cobbe.
\newblock Let's verify step by step, 2023.

\bibitem{Bengio2003}
Yoshua Bengio, R{\'e}jean Ducharme, Pascal Vincent, and Christian Janvin.
\newblock A neural probabilistic language model.
\newblock {\em J. Mach. Learn. Res.}, 3:1137--1155, March 2003.

\bibitem{HochSchm97}
Sepp Hochreiter and J{\"u}rgen Schmidhuber.
\newblock Long short-term memory.
\newblock {\em Neural Computation}, 9(8):1735--1780, 1997.

\bibitem{Vaswani2017attention}
Ashish Vaswani, Noam Shazeer, Niki Parmar, Jakob Uszkoreit, Llion Jones,
  Aidan~N. Gomez, Lukasz Kaiser, and Illia Polosukhin.
\newblock Attention is all you need.
\newblock In I.~Guyon, U.~V. Luxburg, S.~Bengio, H.~Wallach, R.~Fergus,
  S.~Vishwanathan, and R.~Garnett, editors, {\em Advances in Neural Information
  Processing Systems 30}, pages 5998 -- 6008. Curran Associates, Inc., 2017.

\bibitem{Radford2018}
Alec Radford, Jeffrey Wu, Rewon Child, David Luan, Dario Amodei, and Ilya
  Sutskever.
\newblock Language models are unsupervised multitask learners.
\newblock 2018.

\bibitem{Brown2020}
Tom~B. Brown, Benjamin Mann, Nick Ryder, Melanie Subbiah, Jared Kaplan,
  Prafulla Dhariwal, Arvind Neelakantan, Pranav Shyam, Girish Sastry, Amanda
  Askell, Sandhini Agarwal, Ariel Herbert-Voss, Gretchen Krueger, Tom Henighan,
  Rewon Child, Aditya Ramesh, Daniel~M. Ziegler, Jeffrey Wu, Clemens Winter,
  Christopher Hesse, Mark Chen, Eric Sigler, Mateusz Litwin, Scott Gray,
  Benjamin Chess, Jack Clark, Christopher Berner, Sam McCandlish, Alec Radford,
  Ilya Sutskever, and Dario Amodei.
\newblock Language models are few-shot learners, 2020.

\bibitem{Christiano2017}
Paul Christiano, Jan Leike, Tom~B. Brown, Miljan Martic, Shane Legg, and Dario
  Amodei.
\newblock Deep reinforcement learning from human preferences, 2017.

\bibitem{Ziegler2020}
Daniel~M. Ziegler, Nisan Stiennon, Jeffrey Wu, Tom~B. Brown, Alec Radford,
  Dario Amodei, Paul Christiano, and Geoffrey Irving.
\newblock Fine-tuning language models from human preferences, 2020.

\bibitem{SuttonB98}
Richard~S. Sutton and Andrew~G. Barto.
\newblock {\em Reinforcement learning - an introduction}.
\newblock Adaptive computation and machine learning. {MIT} Press, 1998.

\bibitem{Williams92}
R.~J. Williams.
\newblock Simple statistical gradient-following algorithms for connectionist
  reinforcement learning.
\newblock {\em Machine Learning}, 8:229--256, 1992.

\end{thebibliography}

\end{document}